\documentclass[
]{ceurart}

\usepackage{listings}
\usepackage{longtable}
\usepackage{tcolorbox}
\usepackage{booktabs}
\usepackage{array}
\usepackage{caption}
\begin{document}

\copyrightyear{2026}
\copyrightclause{Copyright for this paper by its authors.
  Use permitted under Creative Commons License Attribution 4.0
  International (CC BY 4.0).}

\conference{Joint Proceedings of the AIME 2026 Workshops: 1st International Workshop on Multicentric and Privacy-preserving Learning in Healthcare, Foundation Models for Public Health and Epidemiology: From Promise to Practice, and First International Workshop on Knowledge Graphs for Health, July 10, 2026, Ottawa, Canada}

\title{Can Zero-Shot LLMs Predict Child Malnutrition? A Fairness and Temporal Robustness Study}

\author[1]{Muhammad Ashad Kabir}[%
orcid=0000-0002-6798-6535,
email=akabir@csu.edu.au,
]
\cormark[1]

\author[1]{Md Ahshanul Haque}[%
orcid=0000-0003-3452-8367,
email=mdhaque@csu.edu.au
]
\address[1]{School of Computing, Mathematics and Engineering, Charles Sturt University, Bathurst, NSW 2795, Australia}

\cortext[1]{Corresponding author.}

\begin{abstract}
  Child malnutrition remains a major public health challenge in low- and middle-income countries, particularly in South Asia, where early identification of vulnerable children is critical for timely intervention and resource allocation. This study aims to evaluate the feasibility, fairness, and temporal robustness of using a pretrained large language model (LLM) in a zero-shot setting for child stunting prediction using population health survey data. Using Bangladesh Demographic and Health Survey (BDHS) data collected between 2007 and 2022, we transformed maternal, child, healthcare, and household characteristics into semantically interpretable prompt-based representations and evaluated GPT-4o-mini for zero-shot stunting prediction, comparing its performance against a random forest baseline and assessing fairness across demographic and socioeconomic groups as well as temporal robustness across survey waves. The results demonstrate that zero-shot inference using GPT-4o-mini achieved comparable balanced accuracy to the supervised baseline while exhibiting substantially higher sensitivity for identifying stunting cases, relatively consistent performance across child sex groups, and stable predictive behaviour across BDHS waves; however, important fairness disparities were observed across residence and household wealth categories, highlighting the need for further investigation before deployment of foundation models in public health prediction settings.
\end{abstract}


\maketitle

\section{Introduction}
Child malnutrition remains a major global public health challenge, particularly in low- and middle-income countries, where undernutrition during early childhood is associated with impaired growth, cognitive development, increased disease susceptibility, and elevated mortality risk \cite{victora2008maternal,black2013maternal}. Despite improvements in maternal and child healthcare, South Asian countries, including Bangladesh, continue to experience substantial burdens of childhood stunting, wasting, and underweight \cite{unicef2023malnutrition,bdhs2022}. Early identification of vulnerable children is therefore important for timely intervention and evidence-based public health planning.

Maternal, child, healthcare, and household socioeconomic characteristics have consistently been identified as important determinants of child malnutrition \cite{black2013maternal,victora2008maternal}. Motivated by the increasing availability of Bangladesh Demographic and Health Survey (BDHS) datasets, recent studies have applied conventional machine learning (ML) approaches such as logistic regression, random forests, and gradient boosting to predict childhood malnutrition outcomes \cite{talukder2020machine,khan2021model,islam2024prediction}. However, these studies primarily rely on task-specific supervised learning and provide limited investigation into the applicability of large language models (LLMs) for public health prediction tasks.

Recent advances in LLMs have demonstrated strong zero-shot reasoning capabilities across various biomedical and healthcare applications \cite{bommasani2021opportunities,singhal2023large,thirunavukarasu2023large}. Nevertheless, the use of zero-shot LLMs for structured population health survey data remains largely unexplored, particularly regarding fairness across demographic and socioeconomic groups and temporal robustness.

In this study, we evaluate a pretrained LLM, GPT-4o-mini, in a zero-shot setting for child stunting prediction using BDHS data collected between 2007 and 2022. Using prompt-based feature--value representations of maternal, child, healthcare, and household characteristics, we assess predictive performance, fairness across child sex, place of residence, and household wealth categories, and temporal robustness across BDHS waves. Our findings contribute to the emerging discussion on the practical applicability and limitations of LLMs for population-level public health prediction tasks.

\section{Methodology}

\subsection{Dataset}
This study utilised child-level BDHS data collected between 2007 and 2022 \cite{bdhs2022,dhsprogram}. BDHS is a nationally representative cross-sectional survey containing extensive maternal, child, healthcare, demographic, and household socioeconomic information, and has been widely used in population health and machine learning studies \cite{talukder2020machine,khan2021model,islam2024prediction}. Children under five years of age with complete anthropometric and covariate information were included in the analysis. The target outcome was childhood stunting, defined according to the World Health Organization (WHO) child growth standards as height-for-age z-score (HAZ) below $-2$ standard deviations from the WHO reference population \cite{who2006child}. Features considered in this study were identified as significant determinants of child malnutrition in prior literature \cite{black2013maternal,victora2008maternal,talukder2020machine,islam2024prediction} and are summarised in Table~\ref{tab:dataset_summary}. Samples with missing values for any selected feature were excluded during preprocessing, resulting in a final analytical dataset consisting of 17106 children, including 5623 stunted and 11483 non-stunted cases.

\begin{table}[h]
\centering
\caption{Summary of the selected features used in this study stratified by child stunting status. Values are presented as mean $\pm$ standard deviation or frequency (\%). For binary variables, only one category (e.g., Yes or No) is reported for brevity. The $p$-values indicate statistical differences between stunted and non-stunted groups.}
\label{tab:dataset_summary}
\end{table}
\vspace{-1.5em}
\begin{longtable}{p{4.7cm} p{2.6cm} p{2.6cm} p{2.6cm} p{1.8cm}}
\toprule
Feature name & Feature value & \makecell{Stunting\\(n=5623, \%)} & \makecell{Non-stunting\\(n=11483, \%)} & \textit{p}-value \\
\midrule
\endfirsthead

% \caption[]{Dataset summary (continued).}\\
\multicolumn{5}{l}{\small{Table 1: Dataset summary continued.}} \\
\toprule
Feature name & Feature value &  \makecell{Stunting\\(n=5623, \%)} & \makecell{Non-stunting\\(n=11483, \%)} & \textit{p}-value \\
\midrule
\endhead

\midrule
\multicolumn{5}{r}{\textit{Continued on next page}}\\
\endfoot

\bottomrule
\endlastfoot

Mother's age (years) & Mean $\pm$ SD & 25.37 $\pm$ 6.07 & 25.16 $\pm$ 5.67 & $p=0.032$ \\

Mother's education level & Secondary & 2449 (43.6) & 5674 (49.4) & $p<0.001$ \\
 & Primary & 1838 (32.7) & 2787 (24.3) & \\
 & Tertiary & 384 (6.8) & 1970 (17.2) & \\
 & No education & 952 (16.9) & 1052 (9.2) & \\

Mother's BMI & Mean $\pm$ SD & 20.82 $\pm$ 3.59 & 21.96 $\pm$ 3.91 & $p<0.001$ \\

Worked in last 12 months & No & 4290 (76.3) & 8868 (77.2) & $p=0.176$ \\

Number of living children & Mean $\pm$ SD & 2.23 $\pm$ 1.35 & 2.01 $\pm$ 1.16 & $p<0.001$ \\

Mother's age at first birth & Mean $\pm$ SD & 17.98 $\pm$ 3.09 & 18.69 $\pm$ 3.45 & $p<0.001$ \\

Involved in decisions & Yes & 2558 (45.5) & 5461 (47.6) & $p=0.011$ \\

Believes wife beating & No & 3997 (71.1) & 8818 (76.8) & $p<0.001$ \\

Exposed to mass media & Yes & 3165 (56.3) & 7693 (67.0) & $p<0.001$ \\

Experienced death of child & No & 4848 (86.2) & 10404 (90.6) & $p<0.001$ \\

Father's education level & Secondary & 1598 (28.4) & 3914 (34.1) & $p<0.001$ \\
 & Primary & 1972 (35.1) & 3169 (27.6) & \\
 & No education & 1546 (27.5) & 1940 (16.9) & \\
 & Tertiary & 507 (9.0) & 2460 (21.4) & \\

Father's occupation & Worker & 2628 (46.7) & 5157 (44.9) & $p<0.001$ \\
 & Business or prof & 1322 (23.5) & 3708 (32.3) & \\
 & Agriculture & 1563 (27.8) & 2375 (20.7) & \\
 & Not working & 110 (2.0) & 243 (2.1) & \\

Child age (months) & Mean $\pm$ SD & 24.46 $\pm$ 13.69 & 19.37 $\pm$ 14.45 & $p<0.001$ \\

Child sex & Male & 3014 (53.6) & 5901 (51.4) & $p=0.007$ \\

Birth order of the child & Mean $\pm$ SD & 2.41 $\pm$ 1.55 & 2.12 $\pm$ 1.30 & $p<0.001$ \\

Birth interval & $\geq$24 months & 3238 (57.6) & 6360 (55.4) & $p<0.001$ \\
 & No prior birth & 1904 (33.9) & 4457 (38.8) & \\
 & $<$24 months & 481 (8.6) & 666 (5.8) & \\

Breastfeeding started early & Yes & 4212 (74.9) & 8201 (71.4) & $p<0.001$ \\

Child is being breastfed & Yes & 4205 (74.8) & 9028 (78.6) & $p<0.001$ \\

\makecell[t l]{Child received vitamin A suppl} & Yes & 3614 (64.3) & 6829 (59.5) & $p<0.001$ \\

Skilled birth attendant & No & 3602 (64.1) & 5389 (46.9) & $p<0.001$ \\

Delivered in a health facility & No & 3817 (67.9) & 5909 (51.5) & $p<0.001$ \\

Delivery by caesarean section & No & 4623 (82.2) & 7924 (69.0) & $p<0.001$ \\

\makecell[t l]{Received at least 4 ANC visits} & No & 4284 (76.2) & 7321 (63.8) & $p<0.001$ \\

Drinking water source & Safe & 4943 (87.9) & 9889 (86.1) & $p=0.001$ \\

Type of toilet facility & Hygienic & 2883 (51.3) & 7030 (61.2) & $p<0.001$ \\

Place of residence was rural & Yes & 4103 (73.0) & 7575 (66.0) & $p<0.001$ \\

Household wealth index & Poorest & 1621 (28.8) & 1947 (17.0) & $p<0.001$ \\
 & Richest & 679 (12.1) & 2787 (24.3) & \\
 & Richer & 961 (17.1) & 2465 (21.5) & \\
 & Poorer & 1261 (22.4) & 2086 (18.2) & \\
 & Middle & 1101 (19.6) & 2198 (19.1) & \\

Number of household & Mean $\pm$ SD & 6.01 $\pm$ 2.68 & 6.12 $\pm$ 2.71 & $p=0.011$ \\

\end{longtable}

\subsection{Prompt Construction and Zero-Shot Inference}
Each child's record was represented by serializing feature--value pairs into a structured list-based prompt format (Table~\ref{tab:prompt_template}). Prior studies have shown that list-style serialization preserves tabular semantics and reduces ambiguity when applying LLMs to structured data inference tasks \cite{hegselmann2023tabllm}. Following instruction-tuning paradigms used in modern LLMs, the prompt explicitly defined the task objective, binary output space, and serialized input record. The prompt template was iteratively refined through empirical validation following prior instruction-based prompt optimization approaches.
\begin{table}[h]
\centering
\caption{Example of the prompt used for zero-shot inference.}
\label{tab:prompt_template}

\begin{tcolorbox}[
    colback=gray!8,
    colframe=black,
    boxrule=0.6pt,
    arc=2pt,
    left=6pt,
    right=6pt,
    top=6pt,
    bottom=6pt,
    width=0.95\linewidth
]
\small
You are a clinical classification model. Based on the provided characteristics,
predict whether the child is likely to be stunted. Output must be a single token:
0 = not stunted, 1 = stunted.

\medskip

\noindent
Characteristics: Mother's current age (years): 24, Child sex: Female, ...

\medskip

\noindent
Answer with exactly one token (0 or 1).
\end{tcolorbox}

\end{table}

GPT-4o-mini was used as a pretrained LLM for zero-shot inference due to its strong instruction-following capability, cost efficiency, and support for token-level log-probabilities, enabling probabilistic classification through likelihood comparison of predefined output tokens. Inference through the OpenAI API was performed using deterministic decoding with a temperature set to 0. For baseline comparison, a random forest classifier with \texttt{class\_weight=`balanced'} and \texttt{random\_state=42} was implemented to account for class imbalance.

\section{Results and Discussion}
Table~\ref{tab:subgroup_temporal_performance} presents the fairness and temporal robustness performance of the zero-shot GPT-4o-mini model for child stunting prediction. Overall, GPT-4o-mini achieved a balanced accuracy of 58\%, and AUROC of 0.632, which were reasonably comparable to the random forest baseline model evaluated using 5-fold cross-validation (balanced accuracy: 57\%, AUROC: 0.685). Notably, GPT-4o-mini demonstrated substantially higher sensitivity (77.5\%) than the random forest (23.4\%), indicating greater capability in identifying stunting cases. However, this was accompanied by considerably lower specificity (38.6\% versus 90.7\%), suggesting that the zero-shot LLM tended to over-predict positive stunting outcomes relative to the supervised machine learning baseline.

\begin{table}[ht]
\centering
\caption{Fairness and temporal robustness performance of the zero-shot LLM for child stunting prediction. CV: cross-validation.}
\label{tab:subgroup_temporal_performance}

\begin{tabular}{llrcccc}
\toprule
Category & Value & Test & Balanced & AUROC & Sensitivity & Specificity \\
 & & sample & accuracy (\%) &  & (\%) & (\%) \\
\midrule
\multicolumn{2}{l}{Random Forest} & 5-fold CV & 57.0 & 0.685 & 23.4 & 90.7\\ 
GPT-4o-mini &         &17106 & 58.0 & 0.632 & 77.5 & 38.6 \\
\midrule
\multicolumn{7}{l}{\textit{Fairness of GPT-4o-mini}} \\
\midrule
Child sex & Male  & 8915 & 58.1 & 0.631 & 76.7 & 39.4 \\
          & Female & 8191 & 58.0 & 0.633 & 78.3 & 37.7 \\
\midrule
Residence & Rural & 11678 & 55.4 & 0.620 & 86.0 & 24.9 \\
          & Urban & 5428 & 59.8 & 0.645 & 54.4 & 65.2 \\
\midrule
Wealth  & Poorest & 3568 & 50.0 & 0.581 & 100.0 & 0.0 \\
index    & Poorer & 3347 & 50.9 & 0.560 & 98.7  & 3.0 \\
         & Middle & 3299 & 53.1 & 0.563 & 73.0  & 33.1 \\
             & Richer & 3426 & 55.0 & 0.578 & 54.1  & 55.9 \\
             & Richest & 3466 & 52.9 & 0.583 & 24.7  & 81.1 \\
\midrule
\multicolumn{7}{l}{\textit{Temporal robustness of GPT-4o-mini}}\\
\midrule
BDHS       & 2007 & 219 & 58.2 & 0.611 & 47.6 & 68.9 \\
round      & 2011 & 2587 & 56.5 & 0.622 & 83.8 & 29.2 \\
           & 2014 & 1661 & 56.6 & 0.621 & 77.1 & 35.9 \\
           & 2018 & 1676 & 57.9 & 0.617 & 73.6 & 42.3 \\
           & 2022 & 909 & 57.9 & 0.618 & 71.7 & 44.1 \\
\bottomrule
\end{tabular}
\end{table}

The fairness evaluation showed relatively consistent performance across child sex groups, with nearly identical balanced accuracy and AUROC values for male and female children. In contrast, larger disparities were observed across residence and wealth categories. GPT-4o-mini exhibited substantially higher sensitivity for rural children (86\%) but lower specificity (24.9\%), indicating a tendency to infer stunting more frequently among rural populations. Similarly, the model achieved 100\% sensitivity and 0\% specificity for the poorest wealth category, suggesting potential over-classification of stunting among socioeconomically disadvantaged children. To further investigate whether this behaviour was driven primarily by the wealth index variable itself, we re-evaluated the 3568 samples from the poorest category after excluding the wealth index feature from the prompt. The model still demonstrated very high sensitivity (96\%) and low specificity (6\%), suggesting that the observed prediction pattern may not solely originate from explicit wealth information, but potentially from correlated maternal, household, or other characteristics. These findings warrant further investigation regarding implicit socioeconomic associations learned by pretrained LLMs.

The temporal robustness analysis demonstrated relatively stable performance across BDHS survey waves between 2007 and 2022. Balanced accuracy remained within a narrow range (56.5\%--58.2\%), while AUROC values varied only modestly between 0.611 and 0.622. These findings suggest that the zero-shot LLM maintained reasonably consistent predictive behaviour despite evolving demographic, socioeconomic, and healthcare distributions across survey years. Nevertheless, sensitivity and specificity varied substantially across BDHS rounds. Earlier survey waves, particularly 2007, showed lower sensitivity and higher specificity, whereas later waves exhibited the opposite trend. This shift may reflect temporal changes in population characteristics and healthcare access patterns. %captured within the BDHS data

\section{Conclusions and Future Work}
This study evaluated the feasibility of using a pretrained LLM, GPT-4o-mini, in a zero-shot setting to predict child stunting. The findings based on BDHS dataset demonstrate that zero-shot inference using a pretrained LLM can achieve comparable balanced accuracy to a conventional random forest baseline while exhibiting substantially higher sensitivity for identifying stunting cases. The fairness analysis revealed relatively consistent performance across child sex but highlighted important disparities across residence and household wealth categories, indicating the need for further investigation on potential socioeconomic biases in LLM-based public health prediction. The temporal robustness evaluation further showed relatively stable predictive performance across BDHS waves despite changing population distributions over time. Overall, the study highlights both the potential and limitations of LLMs for population-level health prediction tasks without task-specific training. Future work should further investigate fairness implications, explore few-shot prompting and prompt optimization strategies, and compare different LLMs with robust ML approaches to improve the reliability and generalizability of foundation models for public health applications.

%% The declaration on generative AI comes in effect
%% https://ceur-ws.org/GenAI/Policy.html
\section*{Declaration on Generative AI}
 During the preparation of this work, the authors used ChatGPT-5.4 for grammar and spelling checks. After using this tool, the authors reviewed and edited the content as needed and take full responsibility for the publication's content. 

%%
%% Define the bibliography file to be used
\bibliography{references}

\end{document}